\documentclass[letterpaper, 10 pt, conference]{ieeeconf}
\usepackage{amsmath}
\usepackage{amssymb}
\usepackage{adjustbox}
\usepackage{subcaption}
\usepackage{caption}
\usepackage{calc}
\usepackage{booktabs}
\usepackage{colortbl}
\usepackage{tabularx}
\usepackage{tabu}
\usepackage{multirow}
\usepackage{array}
\usepackage[table]{xcolor}
\usepackage{tikz,pgfplots}
\usepackage[breaklinks=true,bookmarks=false,colorlinks=true]{hyperref} %
\usepackage{amssymb}%
\usepackage{pifont}%
\usepackage{xspace}
\usepackage{cite}
\usepackage{graphicx}

\usetikzlibrary{arrows.meta, positioning, calc, shadows}

\newcommand{\parag}[1]{\vskip2pt \noindent \textbf{#1}}

\newcommand{\cmark}{\ding{51}}%
\newcommand{\xmark}{\ding{55}}%
\definecolor{arrowgray}{RGB}{150,150,150}
\definecolor{profgreen}{RGB}{0,128,0}
\definecolor{darkgreen}{RGB}{0,255,0}
\definecolor{brightpurple}{RGB}{220,110,200}

\newlength{\myXheight}
\definecolor{m_red}{RGB}{255,209,209}
\definecolor{m_red_border}{RGB}{215,23,20}

\definecolor{m_orange}{RGB}{226,213,231}
\definecolor{m_orange_border}{RGB}{150,114,164}

\definecolor{m_blue}{RGB}{217,232,251}
\definecolor{m_blue_border}{RGB}{107,141,190}

\definecolor{m_yellow}{RGB}{255,242,205}
\definecolor{m_yellow_border}{RGB}{213,182,82}

\definecolor{m_gray}{RGB}{245,245,245}
\definecolor{m_gray_border}{RGB}{102,102,102}

\DeclareRobustCommand{\colorsquare}[1]{\tikz{\path[draw=#1_border,fill=#1, thick, rounded corners=0.6pt] (0,0) rectangle (6pt,6pt);}}

\DeclareRobustCommand{\colordot}[2][brightpurple]{\tikz{\filldraw[color=#1, fill=#2, thick](0,0) circle (2pt);}}

\def\eg{\emph{e.g.}\@\xspace} 
\def\ie{\emph{i.e.}\@\xspace}

\def\etal{\emph{et al.}\@\xspace}

\newcommand{\circlenum}[1]{{\textcircled{\scriptsize{#1}}}}

\IEEEoverridecommandlockouts
\title{\LARGE \bf
Mask4Former: Mask Transformer for 4D Panoptic Segmentation
}

\author{Kadir Yilmaz$^{1}$, Jonas Schult$^{1}$, Alexey Nekrasov$^{1}$, Bastian Leibe$^{1}$%
\thanks{$^{1}$Computer Vision Group, RWTH Aachen University, Germany.
{\ \ \ \ \ \ \ Project Page: \footnotesize \url{https://vision.rwth-aachen.de/Mask4Former}}
}%
}

\begin{document}

\maketitle
\thispagestyle{empty}
\pagestyle{empty}

\begin{abstract}
Accurately perceiving and tracking instances over time is essential for the decision-making processes of autonomous agents interacting safely in dynamic environments.
With this intention, we propose Mask4Former for the challenging task of 4D panoptic segmentation of LiDAR point clouds.
Mask4Former is the first transformer-based approach unifying semantic instance segmentation and tracking of sparse and irregular sequences of 3D point clouds into a single joint model.
Our model directly predicts semantic instances and their temporal associations without relying on hand-crafted non-learned association strategies such as probabilistic clustering or voting-based center prediction.
Instead, Mask4Former introduces spatio-temporal instance queries that encode the semantic and geometric properties of each semantic tracklet in the sequence.
In an in-depth study, we find that promoting spatially compact instance predictions is critical as spatio-temporal instance queries tend to merge multiple semantically similar instances, even if they are spatially distant.
To this end, we regress 6-DOF bounding box parameters from spatio-temporal instance queries, which are used as an auxiliary task to foster spatially compact predictions.
Mask4Former achieves a new state-of-the-art on the SemanticKITTI test set with a score of 68.4 LSTQ.

\end{abstract}

\section{INTRODUCTION}
LiDAR is a popular sensor modality in the robotics community due to its ability to provide accurate 3D spatial information.
It allows precise scene understanding of the 3D environment over time, which is essential for agents to safely navigate in dynamic environments by predicting traffic movements and identifying potential hazards.
To achieve the full potential of LiDAR data, in this work, we address the task of 4D panoptic segmentation.
That is, given a sequence of LiDAR scans, the goal is to predict the semantic class of each point while consistently tracking object instances.
The research community has made remarkable progress in advancing 3D vision tasks, fueled by the rapid advancement of deep learning methods~\cite{schult2022mask3d,thomas2019kpconv,marcuzzi2023mask} and the availability of large-scale benchmark datasets~\cite{Geiger2012AreWR,caesar2020nuscenes,fong2022panoptic,Sun2019ScalabilityIP}.
Powerful feature extractors~\cite{thomas2019kpconv,choy20194d,zhu2020cylindrical,tang2020searching} that exploit the rich information offered by LiDAR sensors have been proposed, leading to remarkable improvements in object detection~\cite{Shi2019PVRCNNPF, Lang2018PointPillarsFE, Yan2018SECONDSE}, segmentation~\cite{zhu2020cylindrical,thomas2019kpconv,Milioto2019RangeNetF}, and tracking~\cite{Weng20193DMT,Yin2020Centerbased3O}.

To accomplish holistic 3D scene understanding, 4D panoptic segmentation~\cite{aygun20214d} has recently attracted attention.
Traditionally, approaches follow the tracking-by-detection paradigm~\cite{okuma2004boosted} which decouples 4D panoptic segmentation in the subtasks of semantic segmentation~\cite{thomas2019kpconv, Milioto2019RangeNetF}, object detection~\cite{Lang2018PointPillarsFE} and tracking~\cite{Weng20193DMT, Mittal2019JustGW}.
While this separation of segmentation, detection, and tracking allows for independent improvements in each component, it tends to neglect joint learning of temporal relationships with semantic information.
Significant advances in 4D panoptic segmentation methods address this problem by introducing model architectures that approach the task as a whole and predict semantic class labels for each point and temporally consistent instances~\cite{athar20234dformer}.
Recent methods generate instance predictions by grouping proposals in the 4D spatio-temporal volume~\cite{kreuzberg20224d,aygun20214d,hong2021lidar} or learned embedding space~\cite{marcuzzi2022contrastive}.
However, all previous 4D panoptic segmentation methods fundamentally rely on non-learned clustering methods to aggregate tracklets.

\begin{figure}[t]
\centering
\includegraphics[width=\linewidth]{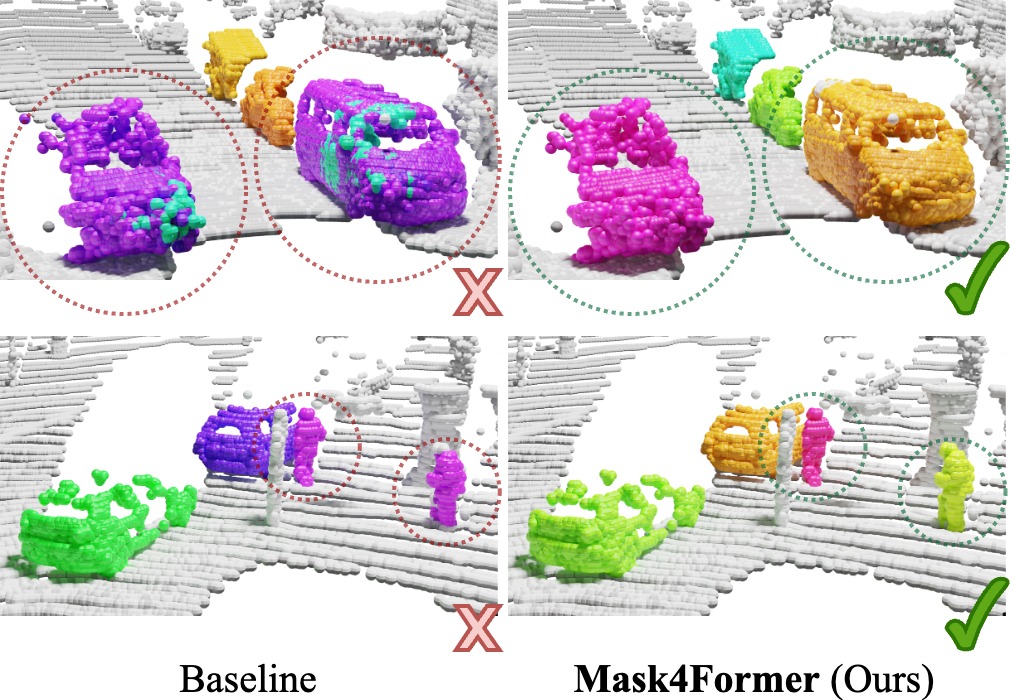}
\caption{\textbf{Spatially non-compact instances.}
Naively adapted for 4D panoptic segmentation, mask transformer approaches reveal a crucial shortcoming: %
instance predictions tend to be spatially non-compact.
As a result, the baseline model predicts two cars as a single object \emph{(left)}.
To overcome this limitation, we introduce Mask4Former, which additionally regresses 6-DOF bounding box parameters for the instance trajectory.
We find that optimizing these bounding box parameters provides a valuable loss signal that promotes spatially compact instances \emph{(right)}.
}
\label{fig:teaser}
\end{figure}

At the same time, we observe a noticeable shift towards unifying tasks~\cite{kirillov2019panoptic, Voigtlaender2019MOTSMT, Yang2019VideoIS} and model architectures~\cite{carion2020end,cheng2021per} for holistic scene understanding.
Central to this trend are mask transformers~\cite{cheng2022masked,schult2022mask3d,cheng2021mask2former} that directly predict foreground masks and their associated semantic labels, eliminating the need for non-learned clustering strategies. %
Typically, these models consist of two main components: a convolutional feature extractor and a transformer decoder.
The convolutional feature extractor processes the point cloud and generates multi-scale features.
The transformer decoder leverages these extracted features and iteratively refines queries each of which encodes the spatial and semantic features for an instance.
Throughout multiple transformer decoder layers, the queries are refined sequentially.
Ultimately, these refined queries directly predict the final semantic class and mask predictions, allowing mask transformers to avoid hand-crafted grouping.
Despite the remarkable performance of mask transformer architectures across diverse tasks, such as image segmentation~\cite{cheng2021per,cheng2022masked}, video segmentation~\cite{cheng2021mask2former}, and 3D scene segmentation~\cite{schult2022mask3d,marcuzzi2023mask,sun22aaai}, it remains open whether such a paradigm generalizes to the unique challenges of 4D panoptic segmentation of sparse point cloud sequences. 

To answer this question, our goal in this paper is to extend mask transformers to 4D panoptic segmentation of point clouds.
Unlike prevailing top-performing approaches for 4D panoptic segmentation~\cite{marcuzzi2022contrastive,kreuzberg20224d,aygun20214d,zhu20234d}, we directly predict foreground masks for \emph{thing} instances and \emph{stuff} regions and their associated semantic labels, bypassing the need for post-processing clustering which requires hand-engineered methods and fine-tuned hyperparameters.
Therefore, in an initial study, we adapt Mask3D~\cite{schult2022mask3d} for 4D panoptic segmentation. %
We follow recent approaches~\cite{aygun20214d,kreuzberg20224d,Hong2022LiDARbased4P} by superimposing consecutive LiDAR scans into a spatio-temporal point cloud that is processed by a sparse convolutional feature backbone~\cite{choy20194d}.
Furthermore, we introduce point-wise spatio-temporal positional encoding in the transformer decoder~\cite{cheng2021mask2former}.
Our findings indicate that these modifications are already competitive with specialized 4D panoptic segmentation methods~\cite{kreuzberg20224d}.
Yet, a deeper examination reveals a significant flaw in mask transformer approaches for 3D point clouds: instances are not always spatially compact~\cite{schult2022mask3d, shen23arxiv}.
Specifically, an instance query may connect multiple instances in the spatio-temporal point cloud, even if they are spatially distant but share semantic similarities (Fig.~\ref{fig:teaser}, \emph{left}).

Based on these findings, we introduce our novel approach called Mask4Former, which is tailored to ensure spatially compact instances, thus unleashing the full potential of mask transformer architectures for 4D panoptic segmentation.
We achieve this by regressing 6-DOF bounding box parameters from the spatio-temporal queries, providing a loss signal to foster spatially compact instance predictions (Fig.~\ref{fig:teaser},~\emph{right}).
We evaluate our Mask4Former model on the challenging SemanticKITTI 4D panoptic segmentation benchmark and achieve state-of-the-art performance %
on the test set. %

In summary, our contributions are fourfold:
\textbf{(1)} We extend the state-of-the-art instance segmentation method Mask3D~\cite{schult2022mask3d} to the 4D panoptic segmentation task.
\textbf{(2)} In experiments, we discover a crucial shortcoming of this straightforward adaptation, namely, the tendency for spatio-temporal instance predictions to lack spatial compactness.
\textbf{(3)} We propose Mask4Former which effectively addresses the aforementioned limitation by introducing a box regression branch that promotes spatially compact instance predictions in an end-to-end trainable fashion, rather than relying on a geometric grouping mechanism with hand-tuned hyperparameters.
\textbf{(4)} Mask4Former achieves state-of-the-art performance on the SemanticKITTI 4D panoptic segmentation benchmark.

\section{RELATED WORK}
\parag{Mask Transformers.}
MaskFormer~\cite{cheng2021per} proposes mask classification as a novel segmentation technique, showcasing its advantages over conventional pixel-based methods.
Inspired by DETR~\cite{carion2020end}, it combines CNNs and transformer networks in a universal segmentation architecture, eliminating the need for task-specific architectures, and streamlining development processes.
Subsequently, Mask2Former~\cite{cheng2022masked} introduces masked attention in the transformer decoder, directing the attention only to relevant parts of the image, and incorporates high-resolution multi-scale features for segmenting smaller objects.
This improves convergence and performance, achieving state-of-the-art results in 2D segmentation tasks~\cite{zhou2017scene, lin2014microsoft, kirillov2019panoptic}.
The paradigm extends to the video instance segmentation~\cite{cheng2021mask2former} task, where Mask2Former effectively addresses temporal consistency, showcasing its universal applicability.
Inspired by its success in 2D, Mask3D~\cite{schult2022mask3d} applies the mask transformer architecture to the 3D domain by leveraging a sparse convolutional backbone~\cite{choy20194d}, and eliminates the need for the predominantly used center-voting and clustering algorithms~\cite{vu2022softgroup,Jiang20CVPR,engelmann20203d}.
For LiDAR panoptic segmentation, MaskPLS~\cite{marcuzzi2023mask} compares mask transformer architectures with adapted semantic segmentation approaches~\cite{tang2020searching, choy20194d, zhu2020cylindrical,campello2013density, comaniciu2002mean, hong2021lidar}, %
demonstrating the superiority of the mask transformer architecture. %

\begin{figure*}[t!]
\centering
\includegraphics[width=\textwidth]{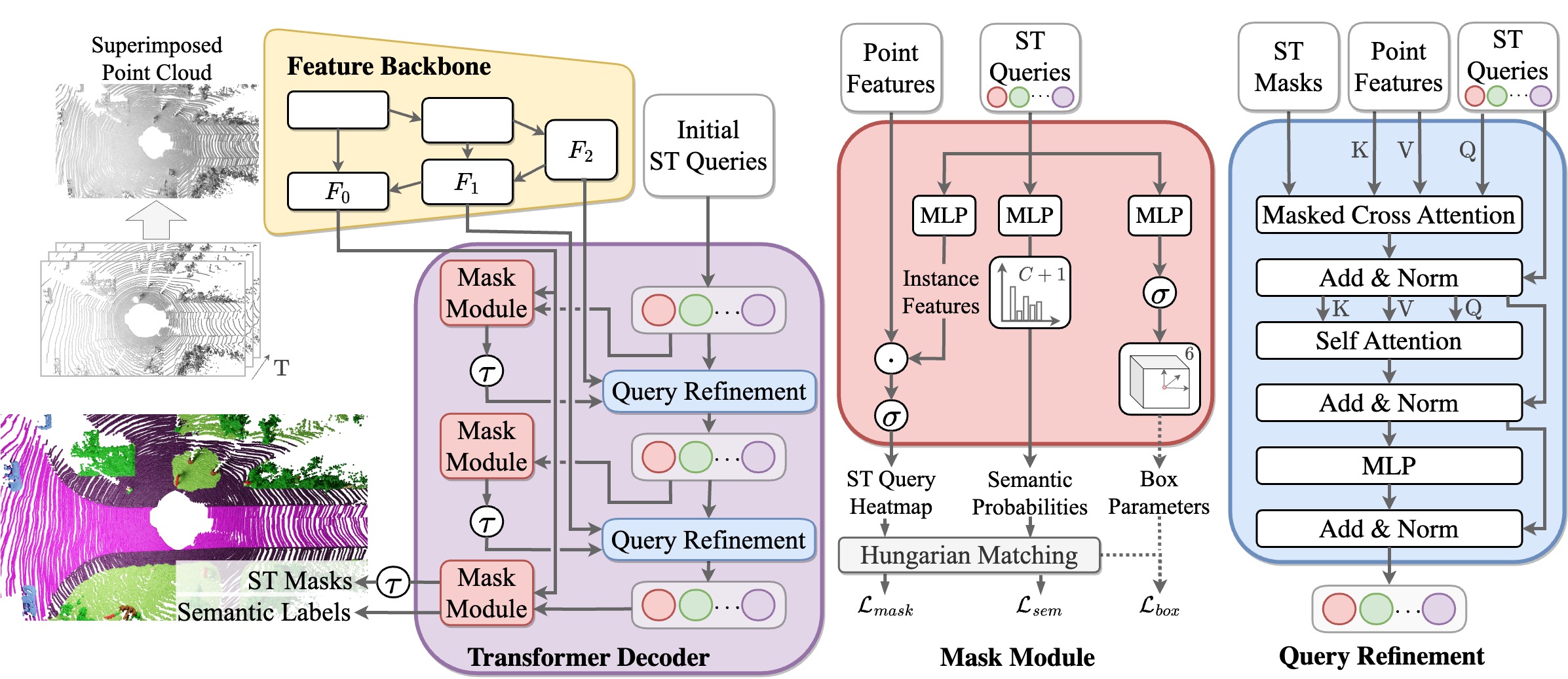}
\caption{\textbf{Illustration of the Mask4Former model.}
We superimpose a sequence of $T$ point clouds into a spatio-temporal representation which is subsequently processed by a sparse convolutional feature backbone~\colorsquare{m_yellow}.
Given a multi-scale feature representation extracted from the feature backbone, the transformer decoder~\colorsquare{m_orange} iteratively refines spatio-temporal (ST) instance queries.
A mask module~\colorsquare{m_red} consumes ST queries and point features at various scales and predicts semantic class probabilities, instance heatmaps, and a 6-DOF bounding box for each ST query.}
\label{fig:model}
\end{figure*}

\parag{4D panoptic segmentation.}
4D-PLS~\cite{aygun20214d} introduces the 4D panoptic segmentation task, associated evaluation metrics, and their method for solving the task.
It superimposes consecutive LiDAR scans to form a spatio-temporal point cloud, performs semantic segmentation, and follows a probabilistic approach for clustering instances based on their predicted centers.
Along the same lines, 4D-DS-Net~\cite{Hong2022LiDARbased4P} and 4D-StOP~\cite{kreuzberg20224d} propose to cluster instances based on spatio-temporal proximity.
4D-DS-Net~\cite{Hong2022LiDARbased4P} extends DS-Net~\cite{hong2021lidar} to the 4D domain by applying a dynamic shifting module to spatio-temporal point clouds which iteratively refines the estimated instance centers and clusters the points in the spatio-temporal volume.
4D-StOP~\cite{kreuzberg20224d}, on the other hand, replaces the probabilistic clustering with an instance-centric voting approach.
Here, initial instance proposals are generated using center votes and then aggregated using learned geometric features.
Building on the success of 4D-StOP, the concurrent work Eq-4D-StOP~\cite{zhu20234d} predicts equivariant fields and incorporates the necessary layers into the models.
This reinforcement of rotation equivariance ensures that the models account for rotational symmetries in the data, resulting in a more robust feature learning.
Contrastingly, CA-Net~\cite{marcuzzi2022contrastive} clusters instances in the feature space.
It leverages an off-the-shelf 3D panoptic segmentation network~\cite{hong2021lidar} and uses extracted point features in a contrastive learning framework~\cite{hadsell2006dimensionality} to generate instance-wise consistent features, resulting in robust instance associations over time. %
Bypassing the need for non-learned clustering approaches, the concurrent work Mask4D~\cite{marcuzzi2023mask4d} adopts the mask transformer-based paradigm but opts for queries that encode single frame instances, and re-uses these queries in subsequent frames to facilitate tracking.
Unlike previous approaches, Mask4Former unifies segmentation and tracking by directly predicting the spatio-temporal instance masks and their corresponding semantic labels. %

\section{Method}
Inspired by the success of mask transformer approaches for 3D instance segmentation~\cite{schult2022mask3d,marcuzzi2023mask,sun22aaai} and 2D video instance segmentation~\cite{cheng2021mask2former}, we propose Mask4Former -- the first mask transformer-based approach for 4D panoptic segmentation.
Building on Mask3D~\cite{schult2022mask3d} for 3D instance segmentation, we introduce technical components that are key to enabling 4D panoptic segmentation of point clouds, \ie, predicting the semantic class of each point and consistently tracking instances over time.

\parag{Overview.}
(Fig.~\ref{fig:model})
As the input to our model, we use a single voxelized point cloud consisting of superimposed consecutive LiDAR scans.
We process the point cloud with a sparse convolutional \emph{feature extractor}~(Fig.\ref{fig:model},~\colorsquare{m_yellow}), which generates a multi-resolution voxel representation for the \emph{transformer decoder}~\colorsquare{m_orange}.
At the core of the model are spatio-temporal (ST) queries that encode geometric and semantic attributes of all instances in a sequence.
To learn ST query features, we use a transformer decoder~\colorsquare{m_orange} that encompasses consecutive query refinement and mask modules.
A \emph{mask module}~\colorsquare{m_red} takes the ST queries and predicts instance heatmaps, semantic class probabilities, and also regresses a bounding box for each instance trajectory.
A \emph{query refinement module}~\colorsquare{m_blue} updates the ST queries by cross-attending to multi-scale voxel representations.
In the following, we provide a detailed description of each component involved.

\parag{Input Spatio-Temporal Point Cloud.}
We represent a temporal sequence of point clouds as a single superimposed and voxelized point cloud.
Similar to other approaches~\cite{aygun20214d,kreuzberg20224d}, we use pose estimates of the ego vehicle~\cite{behley2018efficient,behley2019semantickitti} to create a single scene containing points from multiple LiDAR scans in a global coordinate frame.
Subsequently, this superimposed point cloud represents a spatio-temporal volume, denoted as $\mathcal{P}$~$\in$~$\mathbb{R}^{M\times3}$, which captures the temporal evolution of the scene.
We partition this point cloud into equally sized cubic voxels, thus yielding the representation $\mathcal{V}$~$\in$~$\mathbb{Z}^{K_0 \times 3}$.
This voxelization process not only keeps memory constraints in bounds but also allows for efficient processing of the resulting point cloud by sparse convolutional extractors~\cite{choy20194d}.

\parag{Feature Backbone.}
(Fig.~\ref{fig:model}, \colorsquare{m_yellow})
The sparse convolutional feature extractor processes the voxelized point cloud $\mathcal{V}$~$\in$~$\mathbb{Z}^{K_0 \times 3}$ and extracts multi-scale features $F_r$~$\in$~$\mathbb{R}^{K_r \times D_r}$ at various resolutions $r$.
This design allows the network to capture both local geometry and global context while ensuring the preservation of fine-grained spatial details.

\parag{Mask Module.}
(Fig.~\ref{fig:model}, \colorsquare{m_red})
Each of the $N_q$ ST queries $\mathbf{X}$~$\in$~$\mathbb{R}^{N_q \times D}$ represents a distinct instance over a time period.
The mask module predicts the foreground mask of an instance throughout the sequence and the semantic class of the mask, as well as estimating the 6-DOF bounding box parameters of its trajectory.
To generate this binary foreground mask, ST queries are processed by an MLP, and aligned with the feature space of the backbone's output.
To obtain spatio-temporal masks at the finest resolution, we compute the dot product with the finest backbone features $\mathbf{F}_0$, which -- after sigmoid activation and thresholding -- yields the final binary ST mask.
In addition to these masks, we predict semantic class probabilities for each ST query via a linear projection layer to $C+1$ dimensions, followed by a softmax normalization.
A critical element for consistent tracking of instances over time is the bounding box regression branch.
We feed the ST queries to an MLP followed by sigmoid activation to map the features to a 6-dimensional bounding box parameter space that encodes the normalized bounding box center coordinates $(x,y,z)$ as well as the box dimensions $(w,h,d)$.

\parag{Query Refinement Module.}
(Fig.~\ref{fig:model}, \colorsquare{m_blue})
Following Cheng \etal~\cite{cheng2022masked}, the query refinement blocks refine the ST queries $\mathbf{X}$ by using the voxel features $\mathbf{F}_r$ at various resolutions $r$.
First, a masked cross-attention layer~\cite{cheng2022masked} transforms voxel features $\mathbf{F}_r$ into keys $K$ and values $V$, while ST queries are mapped to queries $Q$.
Here, ST queries attend only to the foreground voxels predicted by the previous mask module. %
We then apply self-attention between queries to ensure that multiple queries do not converge on a single instance.

We use spatio-temporal Fourier positional encodings~\cite{tancik2020fourfeat} to incorporate both spatial and temporal information into our transformer blocks.
To do this, we sum spatial positional encodings based on the voxel positions and temporal positional encodings based on the LiDAR scan time frame~\cite{cheng2021mask2former}.

\parag{Hungarian Matching.}
(Fig.~\ref{fig:model}, \colorsquare{m_gray})
In a single forward pass, Mask4Former determines $N_q$ foreground masks along with their associated semantic class labels.
Since both these predictions and the ground truth targets are not in any particular order, it is necessary to establish optimal one-to-one correspondences between them for model optimization.
Typically, Mask transformer methods~\cite{carion2020end, cheng2021per, cheng2022masked} rely on the Hungarian Algorithm~\cite{Kuhn1955TheHM} for this purpose.
The assignment cost for a predicted semantic mask, \ie, \emph{thing} instances and \emph{stuff} regions, and a target mask is defined as follows:
\begin{equation}
\label{eq:matching}
\mathcal{C}=\mathcal{L}_\text{mask} + \mathcal{L}_\text{sem}
\end{equation}
where $\mathcal{L}_\text{mask}$$=$$\lambda_\text{dice}\mathcal{L}_\text{dice}$$+$$\lambda_\text{BCE}\mathcal{L}_\text{BCE}$ is a weighted combination of the binary cross-entropy loss and the dice loss~\cite{milletari2016fully} for supervising foreground mask predictions and $\mathcal{L}_\text{sem}$$=$$\lambda_\text{CE}\mathcal{L}_\text{CE}$ is the multi-class cross-entropy loss $\mathcal{L}_\text{CE}$ for supervising mask semantics.
The Hungarian algorithm is applied to solve the assignment problem and to find the globally optimal matching that minimizes the total cost while ensuring that each target mask is assigned only once.
The unmatched predicted masks are assigned to a "no-object" mask.

\parag{Training the model.}
After establishing one-to-one correspondences, we can directly optimize each predicted mask.
Our resulting loss consists of three loss functions: We keep the same binary mask loss $\mathcal{L}_\text{mask}$ and the multi-class cross-entropy loss $\mathcal{L}_\text{sem}$ from the Hungarian matching as referenced in Eq.~\ref{eq:matching}.
Observing that the $\mathcal{L}_\text{mask}$ loss does not consider the distance of incorrectly added points to the mask, we introduce a new auxiliary bounding box regression loss $\mathcal{L}_\text{box}$ which promotes spatially compact instances. 
We implement the bounding box loss as an L1 loss on the normalized axis-aligned box parameters. %
By optimizing the bounding box parameters from ST queries, the spatial location of their corresponding masks is supervised.
Consequently, this helps to distinguish similar instances of the same class that are spatially separated.
The overall loss is:
\begin{equation}
\mathcal{L} = \mathcal{L}_\text{mask} + \mathcal{L}_\text{sem} + \mathcal{L}_\text{box}
\end{equation}

\parag{Extracting 4D panoptic segmentations.}
Mask4Former predicts $N_q$ instance tracks as semantic heatmaps which are not necessarily non-overlapping.
To assign a single semantic class label and instance ID to every point within the spatio-temporal point cloud, we proceed in the following manner:
First, for each spatio-temporal query, we obtain semantic confidence by selecting the semantic class with the maximum probability.
Second, this semantic confidence is multiplied with the corresponding instance heatmap, resulting in an overall confidence heatmap.
We then assign each point to the query with the maximum confidence.

\parag{Tracking over long sequences.}
To track instances across long LiDAR sequences that exceed memory limits, it is critical to associate instances across successive spatio-temporal point clouds.
Therefore, we follow Ayg\"un \etal~\cite{aygun20214d} and construct long sequences from short sequences in a way that ensures seamless associations.
We establish a one-to-one match between predicted instances in the last and first frames between short sequences.

\section{EXPERIMENTS}
\subsection{Comparing with State-of-the-Art Methods.}
\parag{Dataset.}
We evaluate Mask4Former on the well-established SemanticKITTI dataset~\cite{behley2019semantickitti}, which is derived from the KITTI odometry dataset~\cite{Geiger2012AreWR}. %
The dataset is split into training, validation, and test sets, and consists of over $43,000$ LiDAR scans recorded with a Velodyne-$64$ laser scanner capturing various urban driving scenarios.
Each point in the LiDAR point clouds is densely annotated with one of $C$$=$$19$ semantic labels, \eg, \emph{car}, \emph{road}, \emph{cyclist}, as well as a unique instance ID that is consistent over time.
For every time step, the dataset includes precise pose estimates of the ego vehicle, which is critical for the 4D panoptic segmentation task.

\parag{Metric.}
The LiDAR Segmentation and Tracking Quality Metric (LSTQ)~\cite{aygun20214d} is designed to evaluate the performance of 4D panoptic segmentation algorithms.
It consists of two main components: classification and association scores.
The classification score $S_{cls}$ evaluates how well the algorithm performs in assigning correct semantic labels to the LiDAR points.
It is calculated as the instance-agnostic mean intersection over union (mIoU) over all classes. %
The association score $S_{assoc}$ evaluates the quality of point-to-instance associations considering the entire LiDAR sequence.
It measures how well the algorithm tracks object instances over time without considering the semantic predictions.
The overall LSTQ metric is computed as the geometric mean of the classification score and the association score: $LSTQ=\sqrt{S_{cls} \times S_{assoc}}$.
The geometric mean ensures that a high score can only be obtained if the approach performs well in both the classification and the association task.
\parag{Implementation Details.}
In all experiments, we use $N_{q}$$=$$100$ ST queries which are initialized with Farthest Point Sampled (FPS) point positions~\cite{schult2022mask3d,qi2017pointnet++}.
Each spatio-temporal point cloud is formed by superimposing 2 consecutive LiDAR scans which are voxelized with a voxel size of $5$~cm.
The sparse feature backbone is a Minkowski Res16UNet34C~\cite{choy20194d}.
We train the model for 30 epochs with a batch size of 4 using the AdamW optimizer~\cite{Loshchilov2017DecoupledWD} and the one-cycle learning rate scheduler~\cite{Smith2017SuperconvergenceVF} with a maximum learning rate of $2\cdot10^{-4}$.
We perform standard data augmentation techniques including random rotation, translation, scaling, and instance population\,\cite{Yan2018SECONDSE}.
For the test set submission, we employ random rotation and translation as test time augmentations to enhance the semantic predictions.

\parag{Results.}
In Tables \ref{table:kitti_test} and \ref{table:kitti_validation}, we report the scores on the SemanticKITTI 4D panoptic segmentation test and validation set, respectively.
Mask4Former outperforms previous approaches by at least +4.5\,LSTQ on the test set and +2.5\,LSTQ on the validation set.
Notably, Mask4Former demonstrates strong semantic understanding by achieving at least +9.0\,S$_\text{cls}$ improvement over previous methods on the test set.

\begin{table}[t]
\setlength{\tabcolsep}{3pt}
\centering
\caption{\textbf{Scores on SemanticKITTI test.}
SemanticKITTI 4D panoptic segmentation test set results.
\emph{Abbreviations:} PP: PointPillars\cite{Lang2018PointPillarsFE}, MOT: Multi-Object Tracking\cite{Weng20193DMT}, SFP: Scene flow based propagation\cite{Mittal2019JustGW}.
$^*$ denotes concurrent work.}
\label{table:kitti_test}
\begin{tabular}{l>{\cellcolor{gray!10}}ccccc}
 \toprule
 Method & LSTQ & S$_\text{assoc}$ & S$_\text{cls}$ & IoU$^{St}$ & IoU$^{Th}$ \\
 \midrule
 KPConv\cite{thomas2019kpconv}+PP+MOT & 38.0 & 25.9 & 55.9 & 66.9 & 47.7 \\
 RangeNet++\cite{Milioto2019RangeNetF}+PP+SFP & 34.9 & 23.3 & 52.4 & 64.5 & 35.8 \\
 KPConv\cite{thomas2019kpconv}+PP+SFP & 38.5 & 26.6 & 55.9 & 66.9 & 47.7 \\
 4D-PLS\cite{aygun20214d} & 56.9 & 56.4 & 57.4 & 66.9 & 51.6 \\
 4D-DS-Net \cite{Hong2022LiDARbased4P} & 62.3 & 65.8 & 58.9 & 65.6 & 49.8 \\
 CIA\cite{marcuzzi2022contrastive} & 63.1 & 65.7 & 60.6 & 66.9 & 52.0 \\
 4D-StOP\cite{kreuzberg20224d} & 63.9 & \underline{69.5} & 58.8 & 67.7 & 53.8 \\
 Mask4D$^*$\cite{marcuzzi2023mask4d} & 64.3 & 66.4 & 62.2 & 69.9 & 52.2 \\
 Eq-4D-StOP$^*$\cite{zhu20234d} & \underline{67.8} & \textbf{72.3} & \underline{63.5} & \underline{70.4} & \underline{61.9} \\
\arrayrulecolor{black!10}\midrule\arrayrulecolor{black}
\textbf{Mask4Former} (Ours) & \textbf{68.4} & 67.3 & \textbf{69.6} & \textbf{72.7} & \textbf{65.3} \\
 \bottomrule
\end{tabular}
\vspace{-10px}
\end{table}

\begin{table}[t]
\centering
\caption{\textbf{Scores on SemanticKITTI validation.}
$^*$ denotes concurrent work.}
\label{table:kitti_validation}
\setlength{\tabcolsep}{3pt}
\begin{tabular}{l>{\cellcolor{gray!10}}ccccc} 
 \toprule
 Method & LSTQ & S$_\text{assoc}$ & S$_\text{cls}$ & IoU$^{St}$ & IoU$^{Th}$ \\
 \midrule
 KPConv\cite{thomas2019kpconv}+PP+MOT & 46.3 & 37.6 & 57.0 & 64.2 & 54.1 \\
 RangeNet++\cite{Milioto2019RangeNetF}+PP+SFP & 43.4 & 35.7 & 52.8 & 60.5 & 42.2 \\
 KPConv\cite{thomas2019kpconv}+PP+SFP & 46.0 & 37.1 & 57.0 & 64.2 & 54.1 \\
 4D-PLS\cite{aygun20214d} & 62.7 & 65.1 & 60.5 & 65.4 & 61.3 \\
 4D-StOP\cite{kreuzberg20224d} & 67.0 & 74.4 & 60.3 & 65.3 & 60.9 \\
 4D-DS-Net\cite{Hong2022LiDARbased4P} & 68.0 & 71.3 & 64.8 & 64.5 & 65.3 \\
 Eq-4D-StOP$^*$\cite{zhu20234d} & 70.1 & \textbf{77.6} & 63.4 & \underline{66.4} & \underline{67.1} \\
 Mask4D$^*$\cite{marcuzzi2023mask4d} & \textbf{71.4} & \underline{75.4} & \textbf{67.5} & 65.8 & \textbf{69.9} \\
\arrayrulecolor{black!10}\midrule\arrayrulecolor{black}
\textbf{Mask4Former} (Ours) & \underline{70.5} & 74.3 & \underline{66.9} & \textbf{67.1} & 66.6 \\
 \bottomrule
\end{tabular}
\end{table}

\subsection{Analysis Experiments.}
\parag{Spatio-Temporal Formation.}
We achieve a globally consistent sequence of LiDAR scans by leveraging the precise pose estimates from the LiDAR sensor~\cite{behley2018efficient}.
Considering that the sparse convolutional feature backbone (Fig.~\ref{fig:model}, \colorsquare{m_yellow}) can process 3- and 4-dimensional inputs~\cite{choy20194d}, we investigate which representation is best for extracting meaningful spatio-temporal features from a sequence.
In Table~\ref{table:ST_formation}, we explore 3 different strategies for representing spatio-temporal feature volumes.
Similar to Cheng~\etal~\cite{cheng2021mask2former}, in the first option \circlenum{1}, we process each LiDAR frame individually and then concatenate them along the spatial dimension before passing them to the Transformer decoder.
In the second option \circlenum{2}, we represent a LiDAR sequence as a 4D feature volume, which is fed into a 4D sparse convolutional feature backbone~\cite{choy20194d}, facilitating the learning of both spatial and temporal relationships directly within the backbone.
Incorporating temporal data early in the backbone shows significant improvements in association quality, yielding an increase of $+6.8$\,S$_\text{assoc}$.
Given the inherent sparsity of point clouds, the third approach \circlenum{3} superimposes, \ie concatenates, several point clouds into a single 3D volume~\cite{aygun20214d,kreuzberg20224d}.
We suspect that superimposing LiDAR scans leads to a denser representation, that is less susceptible to noise, yielding the best performance (Tab.\,\ref{table:ST_formation}).
\begin{table}[t]
\centering
\caption{\textbf{Spatio-Temporal Formation.}
We compare 3 different strategies for representing LiDAR point cloud sequences.
We observe that it is key to enable the feature backbone to incorporate temporal information in the feature representation by creating a 4D spatio-temporal representation or superimposing 3D scans, leading to association improvements of up to $+7.8$\,S$_\text{assoc}$.}
\label{table:ST_formation}
\setlength{\tabcolsep}{5pt}
\begin{tabular}{l>{\cellcolor{gray!10}}ccccc} 
 \toprule
 Feature Extraction & LSTQ & S$_\text{assoc}$ & S$_\text{cls}$ & IoU$^{St}$ & IoU$^{Th}$ \\
 \midrule
\circlenum{1} Sequential 3D & 64.3 & 65.8 & 62.8 & 64.0 & 61.2 \\
\circlenum{2} Spatio-temporal 4D & 68.8 & 72.6 & 65.2 & 66.0 & 64.1 \\
 \arrayrulecolor{black!10}\midrule\arrayrulecolor{black}
\circlenum{3} Superimposed 3D & \textbf{70.2} & \textbf{73.6} & \textbf{66.9} & \textbf{67.2} & \textbf{66.5} \\
 \bottomrule
\end{tabular}
\end{table}

\begin{table}[t]
\centering
\caption{\textbf{Ablation study on bounding box regression.}
We observe that optimizing Mask4Former using the regressed bounding box parameters leads to substantially better association scores compared to the baseline ($+3.5$\,S$_\text{assoc}$). 
}
\label{table:ablation_study}
\begin{tabu}{*{2}{c>{\centering\arraybackslash}m{0.7cm}}*{3}{>{\centering\arraybackslash}m{0.7cm}}}
 \toprule
 & $\mathcal{L_\text{box}}$ & DBS & LSTQ & S$_\text{cls}$ & S$_\text{assoc}$\\
\cmidrule(r{4pt}){2-3} \cmidrule(l{4pt}){4-6}
\circlenum{1} &\xmark & \xmark & 68.6 & \textbf{67.3} & \tikz[remember picture] \node[inner sep=0pt,anchor=base] (n1) {70.1};\\
\circlenum{2} &\xmark & \cmark  & 70.1 & \textbf{67.3} & \tikz[remember picture] \node[inner sep=0pt,anchor=base] (n2) {72.8};\\
\circlenum{3}  &\cmark & \xmark & 70.2 & 66.9 & \tikz[remember picture] \node[inner sep=0pt,anchor=base] (n3) {73.6};\\
 \arrayrulecolor{black!10}\midrule\arrayrulecolor{black}
\circlenum{4} &\cmark & \cmark & \textbf{70.5} & 66.9 & \tikz[remember picture] \node[inner sep=0pt,anchor=base] (n4) {\textbf{74.3}};\\
 \bottomrule
\end{tabu}

\begin{tikzpicture}[overlay, remember picture,>=Stealth]
    \draw[->, arrowgray,rounded corners=2pt] (n1.east) -- ++(0.4cm,0) |- (n2.east) node[pos=0.25, fill=white, inner sep=1pt,text=profgreen] {\tiny{+$2.7$}};
    \draw[->, arrowgray,rounded corners=2pt] (n1.east) -- ++(0.7cm,0) |- (n3.east) node[pos=0.35, fill=white, inner sep=1pt,text=profgreen] {\tiny{+$3.5$}};
    \draw[->, arrowgray,rounded corners=2pt] (n1.east) -- ++(1.0cm,0) |- (n4.east) node[pos=0.72, fill=white, inner sep=1pt,text=profgreen] {\tiny{+$4.2$}};
\end{tikzpicture}

\vspace{-10px}
\end{table}

\parag{Spatially non-compact instance predictions.}
Achieving consistent tracking of multiple instances over time in LiDAR sequences is particularly challenging. 
This is due to the sparsity of the point clouds, as well as the occlusions and deformations that instances undergo over time, requiring robust temporal feature learning.
In an initial study, we analyze our baseline method without the bounding box regression branch in the mask module (Fig.\,\ref{fig:model}, \colorsquare{m_red} and Tab.\,\ref{table:ablation_study}, \circlenum{1}), which reveals a crucial shortcoming of applying mask-transformer approaches directly to the task of 4D panoptic segmentation: Instance predictions tend to lack spatial compactness, \ie, the spatio-temporal queries group multiple instances with similar semantics together, even if they are spatially distant (Fig.\,\ref{fig:teaser}, \emph{left}).
To validate this observation, we apply the density-based clustering method, DBSCAN\,\cite{campello2013density}, to each foreground mask prediction.
This separates the instance mask predictions into spatially compact instances.
The impact was noticeable: applying DBSCAN \circlenum{2} to the instance predictions results in a significant improvement of +2.7 S$_\text{assoc}$, confirming our initial findings and supporting our hypothesis.
Anticipating further improvements by replacing DBSCAN with a learned component, we introduce a specialized box regression branch \circlenum{3} which promotes spatial awareness to better separate instances.
This approach outperforms the baseline, both with and without DBSCAN, by a margin of up to $+3.5$\,S$_\text{assoc}$.
Combining the box regression branch with DBSCAN yields our proposed method Mask4Former \circlenum{4}, which not only ensures a strong association between instances ($+4.2$\,S$_\text{assoc}$) but also achieves strong semantic scene understanding, scoring $66.9$\,S$_\text{cls}$ on the SemanticKITTI validation.

\begin{figure}[t]
    \centering
    \begin{subfigure}[b]{\linewidth}
        \centering
        \includegraphics[width=\linewidth]{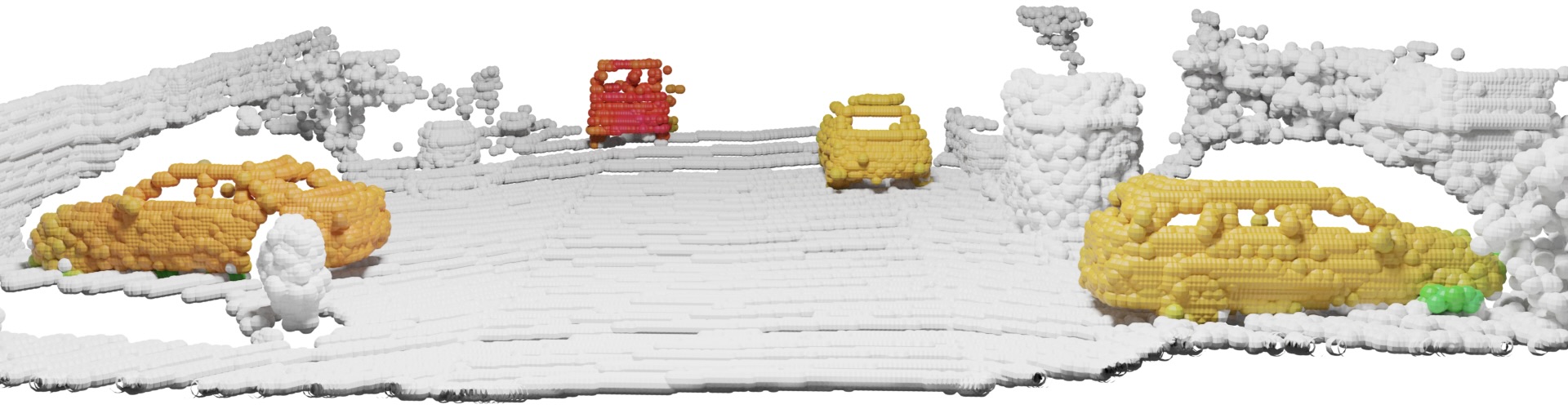}
        \begin{tikzpicture}[overlay, remember picture]
            \coordinate (start) at (8mm, 21mm);  %
            \coordinate (end) at (25mm, 18mm); %
            \node[above, text=red!60!black, font=\Large] at (20mm, 23mm) {\xmark};
            \draw[latex-latex, dashed, very thick, draw=black!90] (start) to[out=30,in=120] (end);
        \end{tikzpicture}
        \vspace{-15pt}
        \caption{Mask4Former without box loss}
        \label{fig:wo_box}
    \end{subfigure}
    
    \vspace{1em}
    
    \begin{subfigure}[b]{\linewidth}
        \centering
        \includegraphics[width=\linewidth]{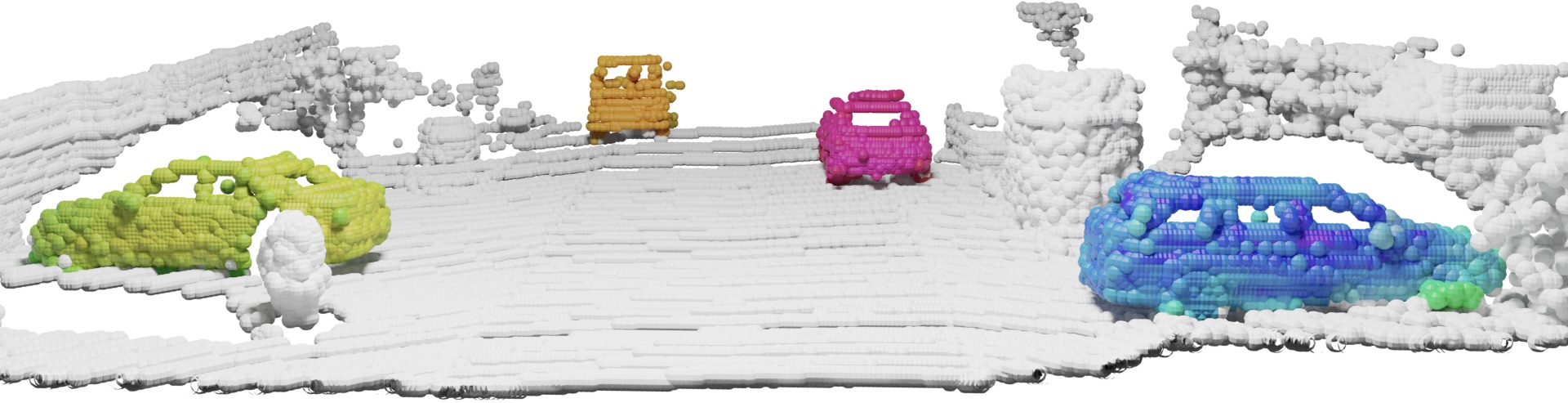}
        \begin{tikzpicture}[overlay, remember picture]
            \coordinate (start) at (8mm, 21mm);  %
            \coordinate (end) at (25mm, 18mm); %
            \node[above, text=green!60!black, font=\Large] at (20mm, 23mm) {\cmark};
            \draw[latex-latex, dashed, very thick, draw=black!90] (start) to[out=30,in=120] (end);
        \end{tikzpicture}
        \vspace{-15pt}
        \caption{\textbf{Mask4Former} (Ours)}
        \label{fig:w_box}
    \end{subfigure}
    \caption{\textbf{Visualization of learned point representations.}
    We use PCA to project the learned point representation of instances into RGB space.
    Our model trained without bounding box supervision, exhibits reduced variance in its feature representation for instances.
    In contrast, Mask4Former effectively separates distinct instances in the feature space.
    \vspace{-13px}
    }
    \label{fig:pca}
\end{figure}
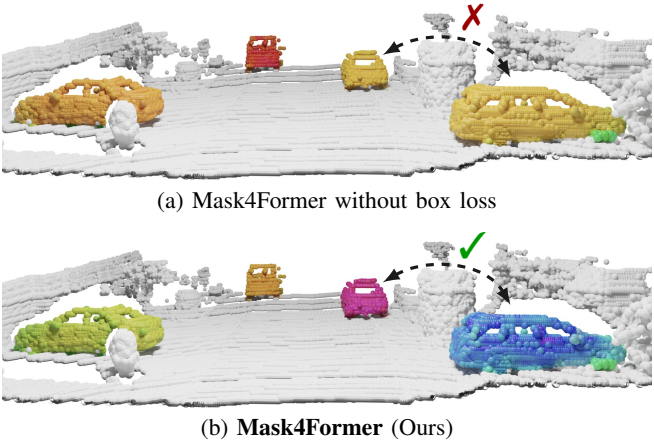

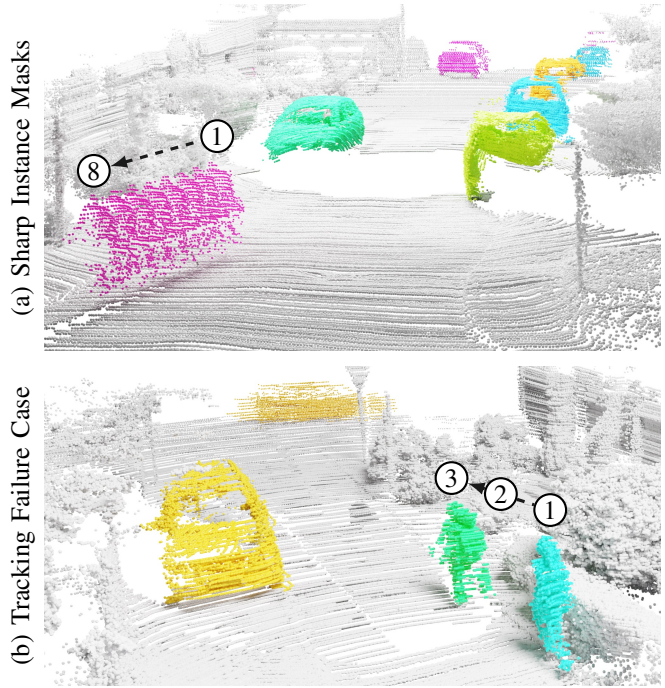
\begin{figure}
    \centering

    \newlength{\captwidth}
    \settoheight{\captwidth}{(a) Sharp Instance Masks}

    \begin{minipage}[c]{\captwidth + 6pt} 
        \rotatebox{90}{\centering (a) Sharp Instance Masks}
    \end{minipage}%
    \begin{minipage}[c]{\linewidth - \captwidth - 6pt} 
    \begin{tikzpicture}
        \node[anchor=south west,inner sep=0] (image) at (0,0) {\includegraphics[width=\linewidth]{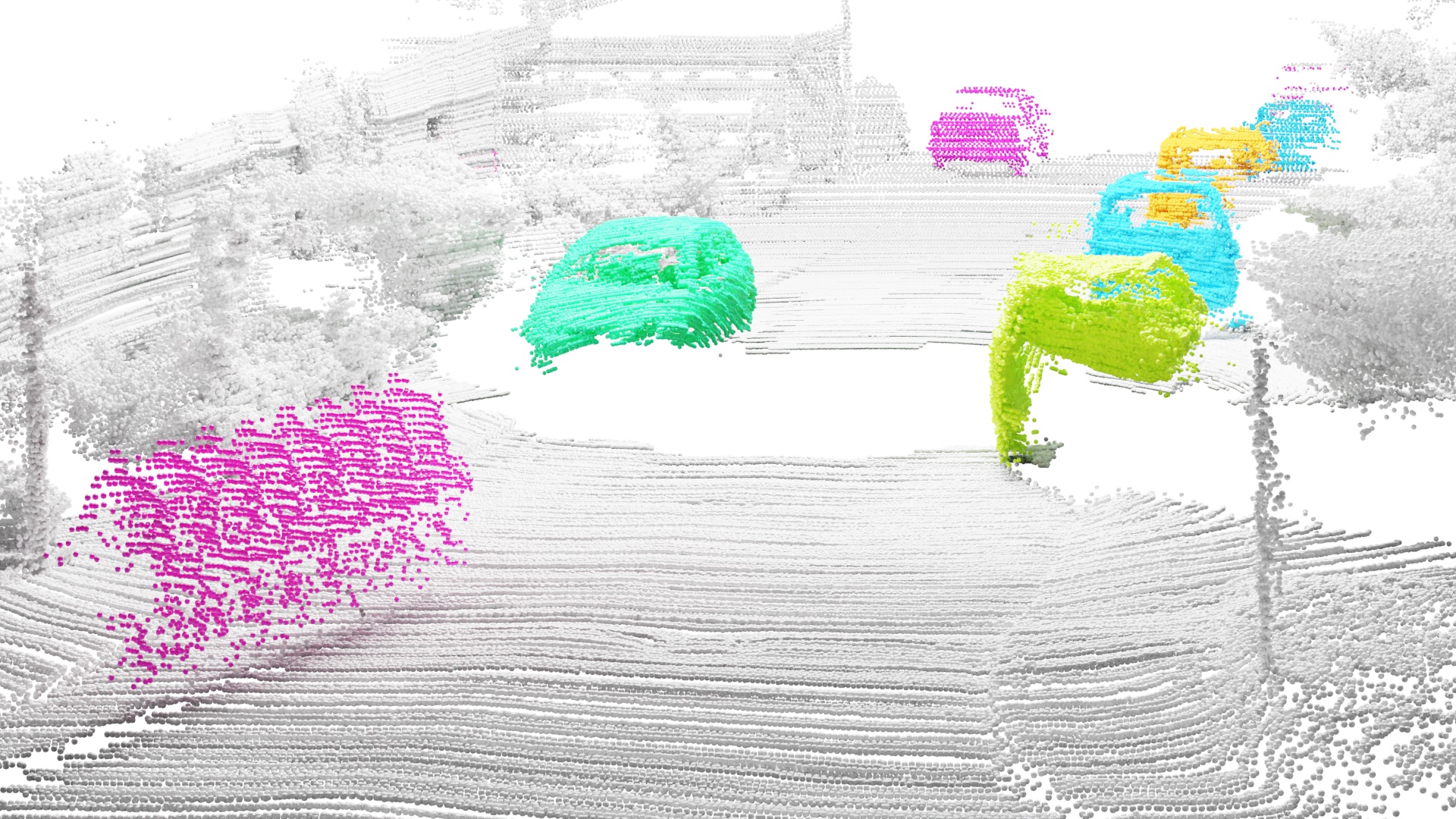}};
        
        \begin{scope}[x={(image.south east)},y={(image.north west)}]

            \tikzstyle{mynode} = [circle, fill=white, fill opacity=1.0, draw, thick, inner sep=1pt, minimum size=12pt]

            \node [mynode] (1) at (0.28, 0.62) {1};
            \node [mynode] (8) at (0.08, 0.52) {8};
            
            \draw [latex-, dashed, very thick, draw=black!90] (8) -- (1);

        \end{scope}
    \end{tikzpicture}
    \end{minipage}

    \vspace{6pt}
    
    \begin{minipage}[c]{\captwidth + 6pt} 
        \rotatebox{90}{\centering (b) Tracking Failure Case}
    \end{minipage}%
    \begin{minipage}[c]{\linewidth - \captwidth - 6pt} 
    \begin{tikzpicture}
        \node[anchor=south west,inner sep=0] (image) at (0,0) {\includegraphics[width=\linewidth, trim=0cm 0cm 0cm 3cm, clip]{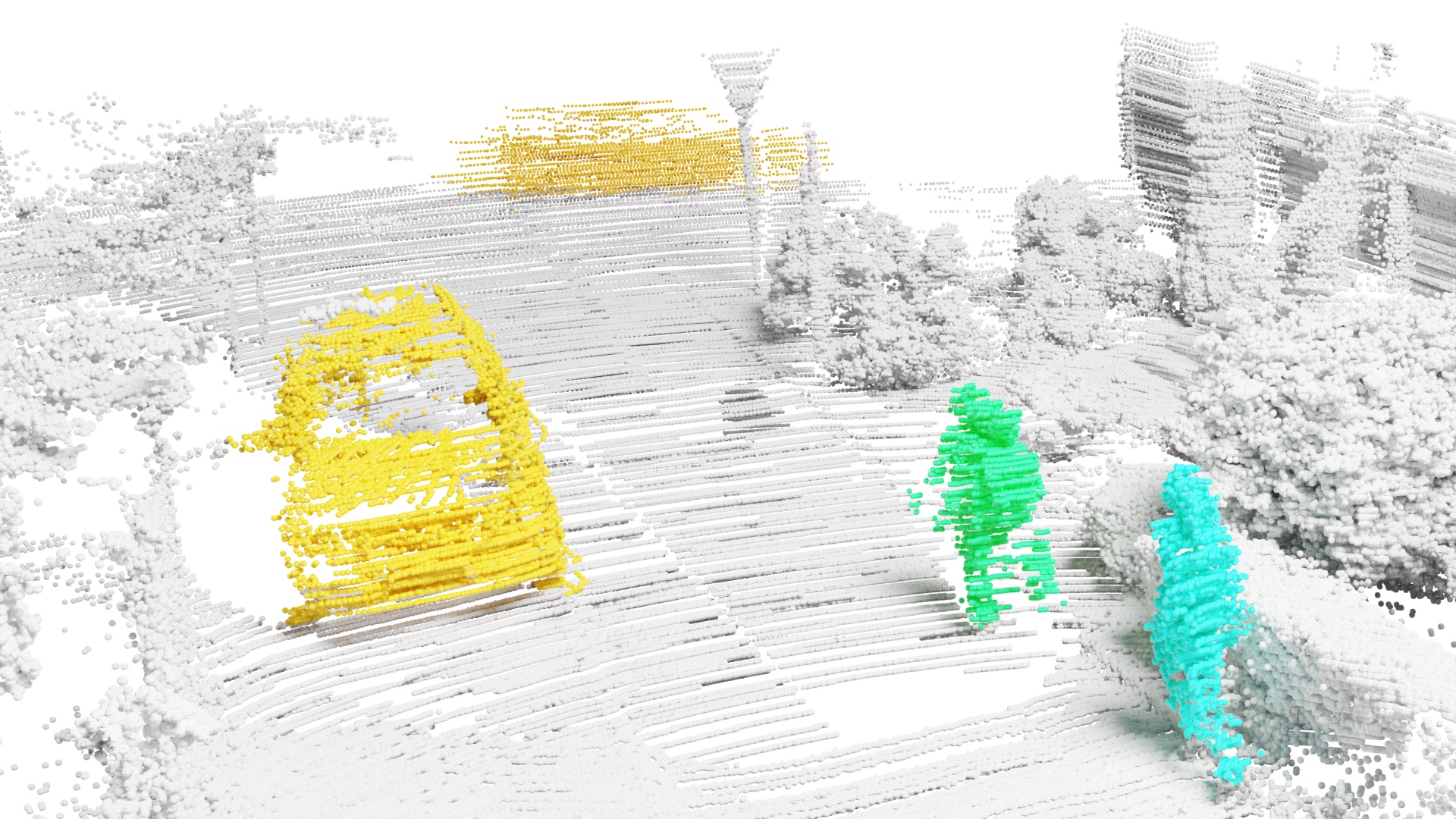}};
        
        \begin{scope}[x={(image.south east)},y={(image.north west)}]

            \tikzstyle{mynode} = [circle, fill=white, fill opacity=1.0, draw, thick, inner sep=1pt, minimum size=12pt]

            \node [mynode] (1) at (0.82, 0.55) {1};
            \node [mynode] (2) at (0.74, 0.60) {2};
            \node [mynode] (3) at (0.66, 0.65) {3};
            
            \draw [latex-, dashed, very thick, draw=black!90] (3) -- (2);
            \draw [-, dashed, very thick, draw=black!90] (2) -- (1);

        \end{scope}
    \end{tikzpicture}
    \end{minipage}
    
\caption{\textbf{Qualitative Results.}
We show color-coded instance tracks over 8 superimposed frames in a spatio-temporal point cloud and a failure where a pedestrian track is split due to an observation being outside of the LiDAR's field of view.
\vspace{-10px}
}
\label{fig:qualitative_results}
\end{figure}

\parag{Visualization of point features learned by Mask4Former.}
In Fig.~\ref{fig:pca}, we show examples of PCA projected features $F_0$ extracted from the finest resolution of Mask4Former's feature backbone
(Fig.\,\ref{fig:model}, \colorsquare{m_yellow}).
When trained without our suggested box loss, Mask4Former shows less distinct separation of instance point features within the feature space (Fig.\,\ref{fig:wo_box}). 
Conversely, the model optimized with the auxiliary task of 6-DOF bounding box regression for each instance trajectory shows a distinct separation of instance point features in the feature space (Fig.\,\ref{fig:w_box}).
This indicates that Mask4Former learns a more semantically meaningful feature space for the task of 4D panoptic segmentation leading to its superior association score $S_\text{assoc}$, as highlighted in Tab.\,\ref{table:ablation_study}.

\parag{Qualitative results.}
In Fig.~\ref{fig:qualitative_results}\textcolor{red}{a}, we show qualitative results.
We observe that Mask4Former not only produces sharp instance masks but also reliably tracks the moving bicyclist\,\colordot{brightpurple} throughout the entire sequence.
We also demonstrate a failure case of our tracking approach.
As we process long sequences by stitching short sequences with overlaps, we incorrectly split tracks when an instance is not present in the overlapping LiDAR scan.
For example, in Fig.~\ref{fig:qualitative_results}\textcolor{red}{b}, a pedestrian near the ego vehicle falls below the LiDAR's field of view.
As a result, when the pedestrian becomes visible again, our tracking approach fails and predicts it as a new instance.

\section{CONCLUSION}
Inspired by the success of recent mask transformer-based approaches, we have extended Mask3D to the task of 4D panoptic segmentation and have achieved promising results.
In an in-depth analysis, we have found that Mask3D for 4D panoptic segmentation tends to produce spatially non-compact instances, resulting in poor association quality.
To overcome this limitation, we have introduced Mask4Former, the first transformer-based approach, that unifies segmentation and tracking of 3D point cloud sequences and is tailored to ensure spatially compact instances.
To this end, Mask4Former regresses 6-DOF bounding box parameters that are optimized to provide a loss signal to encourage spatially compact instance predictions.
Through extensive experimental evaluations, we have demonstrated the effectiveness of Mask4Former, achieving state-of-the-art performance on the SemanticKITTI 4D panoptic segmentation benchmark.
We anticipate follow-up work along the lines of direct prediction of instance and semantic labels.

{\small\parag{Acknowledgments:}
This project is partially funded by the Bosch-RWTH LHC project ``Context Understanding for Autonomous Systems", the BMBF project 6GEM (\verb|16KISK036K|) and the NRW project WestAI (\verb|01IS22094D|).
Compute resources were granted by RWTH Aachen under project \verb|supp0003|.
This work is part of the first author’s master thesis.}

\newpage
\bibliographystyle{plain}
\bibliography{abbrev,egbib}

\clearpage

\twocolumn[{%
\renewcommand\twocolumn[1][]{#1}%
\vspace{0.1cm}
\begin{center}
\textbf{\Large Mask4Former: Mask Transformer for 4D Panoptic Segmentation\\[0.2cm]
\textit{Supplementary Material}}
\end{center}
\vspace{0.5cm}
}]
In this supplementary material, we demonstrate the versatility of Mask4Former by applying it to various segmentation tasks, showcasing its potential as a comprehensive 3D segmentation framework.
Specifically, we use Mask4Former for both 3D panoptic segmentation and 4D semantic segmentation tasks.
Our results indicate that Mask4Former achieves competitive performance across these tasks without any hyperparameter tuning or architectural modifications.

\parag{3D panoptic segmentation} is the task of assigning a semantic class label for each point in a 3D scene while distinguishing different instances of the same class.
Unlike 4D panoptic segmentation, which involves tracking instances over time, 3D panoptic segmentation processes each LiDAR scan independently.
Transitioning from 4D to 3D panoptic segmentation for Mask4Former is straightforward by adjusting the number of superimposed LiDAR scans to 1.
Evaluation is based on the panoptic quality metric,\cite{kirillov2019panoptic}, calculated as follows:
\begin{equation}
\label{eq:PQ}
\mathrm{PQ}=\underbrace{\frac{\sum_{(p, g) \in T P} \operatorname{IoU}(p, g)}{|T P|}}_{\text {segmentation quality (SQ) }} \times \underbrace{\frac{|T P|}{|T P|+\frac{1}{2}|F P|+\frac{1}{2}|F N|}}_{\text {recognition quality (RQ) }}
\end{equation}
For each class, prediction and ground truth masks are sorted into true positives, false positives, and false negatives.
True positives represent pairs of prediction and ground truth masks with an IoU overlap of over 50\%, false positives are unmatched predicted masks, and false negatives are unmatched ground truth masks.
Segmentation quality is determined by the average IoU of matched segments, while recognition quality measures the F1 score, indicating the effectiveness of object recognition.

In Table \ref{table:kitti_panoptic3d_val} we report the scores on the SemanticKITTI 3D panoptic segmentation validation set.
Our model achieves a PQ score of 61.7\%, demonstrating competitive results compared to state-of-the-art techniques without any modifications.
Notably, it outperforms the end-to-end trainable method~\cite{marcuzzi2023mask} by 1.9\%.
Furthermore, our model achieves an SQ score of 80.8\%, indicating that Mask4Former generates precise instance masks.
However, a comparatively lower RQ score suggests that Mask4Former tends to produce many unmatched instance masks.
This discrepancy might stem from the fact that the predicted number of instances ($N_q$) exceeds the actual number of instances in a LiDAR scene, which might result in a multitude of small mask predictions.

\parag{4D semantic segmentation} is a semantic segmentation task where moving and stationary objects of the same category are treated as different semantic classes.
As a result, there are 6 extra classes for moving objects such as "moving car" and "moving person" on top of the regular 19 classes.
To distinguish between moving and stationary objects, the model needs to process multiple LiDAR scans together.
This is the reason why it is also referred to as multi-scan semantic segmentation.
Transitioning from 4D panoptic segmentation to 4D semantic segmentation requires two minor modifications.
Firstly, instead of generating a target mask for each instance, a single target mask per class is generated.
Consequently, the spatiotemporal queries predict all points belonging to a semantic class together.
Secondly, bounding box parameter regression is omitted since a single target mask may encompass multiple instances of the same class.
The evaluation metric for this task is mean Intersection over Union (mIoU) computed across 25 classes.

In Table \ref{table:kitti_sem4d_test} we report the scores on the SemanticKITTI 4D semantic segmentation test set.
Our results demonstrate notable achievements using the Mask4Former framework, with an mIoU of 55.7\%, securing the second position on the leaderboard.
We achieve an mIoU$_\text{static}$ of 59.8\% on par with the state-of-the-art, showing we can accurately segment static objects.
However, the mIoU$_\text{moving}$ of 42.8\% suggests challenges in distinguishing between moving and static objects of the same class.
This discrepancy may be attributed to our decision to superimpose only 2 consecutive LiDAR scans, as opposed to the 13 scans utilized in ~\cite{li2023memoryseg}, to maintain consistency with the 4D panoptic segmentation framework.
This may hinder our model's ability to capture sufficient temporal context for effectively distinguishing between these object categories.

\begin{table}[t]
\centering
\caption{\textbf{3D panoptic segmentation scores on SemanticKITTI validation set.}}
\label{table:kitti_panoptic3d_val}
\setlength{\tabcolsep}{10pt}
\begin{tabular}{l>{\cellcolor{gray!10}}ccc} 
 \toprule
 Method & PQ & SQ & RQ \\
 \midrule
 DS-Net\cite{hong2021lidar}& 57.7 & 77.6 & 68.0 \\
 Panoptic-PolarNet\cite{zhou2021panoptic} & 59.1 & 78.3 & 70.2\\
 EfficientLPS\cite{sirohi2021efficientlps} & 59.2 & 75.0 & 69.8\\
 Mask-PLS\cite{marcuzzi2023mask} & 59.8 & 76.3 & 69.0 \\
 Panoptic-PHNet\cite{li2022panoptic} & 61.7 & - & - \\
 GP-S3Net\cite{razani2021gp} & \underline{63.3} & \underline{81.4} & \textbf{75.9} \\
 PUPS\cite{su2023pups} & \textbf{64.4} & \textbf{81.5} & \underline{74.1} \\
\arrayrulecolor{black!10}\midrule\arrayrulecolor{black}
\textbf{Mask4Former} (Ours) & 61.7 & 81.0 & 71.4 \\
 \bottomrule
\end{tabular}
\end{table}

\begin{table}[t]
\centering
\caption{\textbf{4D semantic segmentation scores on SemanticKITTI test set.}}
\label{table:kitti_sem4d_test}
\setlength{\tabcolsep}{5pt}
\begin{tabular}{l>{\cellcolor{gray!10}}ccc} 
 \toprule
 Method & mIoU & mIoU$_\text{moving}$ & mIoU$_\text{static}$ \\
 \midrule
 TangentConv\cite{tatarchenko2018tangent} & 34.1 & 20.3 & 38.5 \\
 DarkNet53Seg\cite{tatarchenko2018tangent} & 41.6 & 26.3 & 46.4 \\
 SpSequenceNet\cite{shi2020spsequencenet} & 43.1 & 26.5 & 48.3 \\
 TemporalLidarSeg\cite{duerr2020lidar} & 47.0 & 29.8 & 52.4 \\
 TemporalLatticeNet\cite{schutt2022abstract} & 47.1 & 34.5 & 51.1 \\
 Meta-RangeSeg\cite{wang2022meta} & 49.7 & 38.1 & 53.4 \\
 KPConv\cite{thomas2019kpconv}& 51.2 & \underline{43.7} & 53.6 \\
 Cylinder3D\cite{zhu2020cylindrical} & 52.5 & 36.8 & 57.5\\
 MemorySeg\cite{li2023memoryseg} & \textbf{58.3} & \textbf{53.4} & \underline{59.8}\\
\arrayrulecolor{black!10}\midrule\arrayrulecolor{black}
\textbf{Mask4Former} (Ours) & \underline{55.7} & 42.8 & \textbf{59.8} \\
 \bottomrule
\end{tabular}
\end{table}

\end{document}